\documentclass[runningheads]{llncs}

\usepackage{eccv}

\usepackage{eccvabbrv}

\usepackage{graphicx}
\usepackage{booktabs}

\usepackage[accsupp]{axessibility}  

\usepackage[pagebackref,breaklinks,colorlinks,citecolor=eccvblue]{hyperref}

\usepackage{graphicx}
\usepackage{url}
\usepackage{algorithm}
\usepackage{multirow}
\usepackage{enumitem}
\usepackage{algpseudocode}
\hypersetup{colorlinks,allcolors=blue}
\usepackage{amsmath}
\usepackage[capitalize]{cleveref}
\usepackage{appendix}
\usepackage{float}
\usepackage{bbding}
\usepackage{wrapfig} 
\usepackage{booktabs}
\usepackage[most]{tcolorbox}
\usepackage[table,xcdraw]{xcolor}
\usepackage{orcidlink}
\usepackage{xcolor}
\usepackage{mdframed}
\usepackage{fvextra}
\usepackage{tcolorbox}
\tcbset{
  promptbox/.style={
    colback=gray!10,
    colframe=black,
    arc=2mm,
    boxrule=0.8pt,
    breakable,
    left=6pt,
    right=6pt,
    top=6pt,
    bottom=6pt
  }
}
\tcbuselibrary{breakable}
\DefineVerbatimEnvironment{PromptVerb}{Verbatim}{
  breaklines=true,
  breakanywhere=false,
  fontsize=\footnotesize,
  baselinestretch=1,
  commandchars=\\\{\} 
}
\begin{document}

\title{IMTBench: A Multi-Scenario Cross-Modal Collaborative Evaluation Benchmark for \\
In-Image Machine Translation} 

\titlerunning{IMTBench: Multi-Scenario Cross-Modal Collaborative Benchmark}

\author{Jiahao Lyu\inst{1,2,6} \and
Pei Fu\inst{2}* \and 
Zhenhang Li\inst{4} \and
Weichao Zeng\inst{5} \and
Shaojie Zhang\inst{2} \and
Jiahui Yang\inst{2} \and
Can Ma\inst{1} \and
Yu Zhou\inst{3}\textdagger \and
Zhenbo Luo\inst{2} \and
Jian Luan\inst{2}
}

\authorrunning{J.Lyu et al.}

\institute{Institute of Information Engineering, Chinese Academy of Science, China 
\\
\and
MiLM Plus, Xiaomi Inc, China
\\
\and
VCIP $\&$ TMCC $\&$ DISSec, College of Computer Science $\&$ College of Cryptology and Cyber Science, Nankai University, China
\and
Binghamton University, USA
\and
The University of Tokyo, Japan
\\
\and School of Cyber Security, University of Chinese Academy of Science, China 
}

\maketitle

\begin{abstract}
End-to-end In-Image Machine Translation (IIMT) aims to convert text embedded within an image into a target language while preserving the original visual context, layout, and rendering style. However, existing IIMT benchmarks are largely synthetic and thus fail to reflect real-world complexity, while current evaluation protocols focus on single-modality metrics and overlook cross-modal faithfulness between rendered text and model outputs. To address these shortcomings, we present In-image Machine Translation Benchmark (IMTBench), a new benchmark of 2,500 image translation samples covering four practical scenarios and nine languages. IMTBench supports multi-aspect evaluation, including translation quality, background preservation, overall image quality, and a cross-modal alignment score that measures consistency between the translated text produced by the model and the text rendered in the translated image. We benchmark strong commercial cascade systems, and both closed- and open-source unified multi-modal models, and observe large performance gaps across scenarios and languages, especially on natural scenes and resource-limited languages, highlighting substantial headroom for end-to-end image text translation. We hope IMTBench establishes a standardized benchmark to accelerate progress in this emerging task.
  \vspace{-20pt}
\end{abstract}

\let\thefootnote\relax
\footnotetext{Work done when Jiahao Lyu was an intern at Xiaomi. *Project Leader. \textdagger~Corresponding Author.} 

\section{Introduction}
\label{sec:intro}

\begin{figure}[t]
\centering
\includegraphics[width=0.9\textwidth]{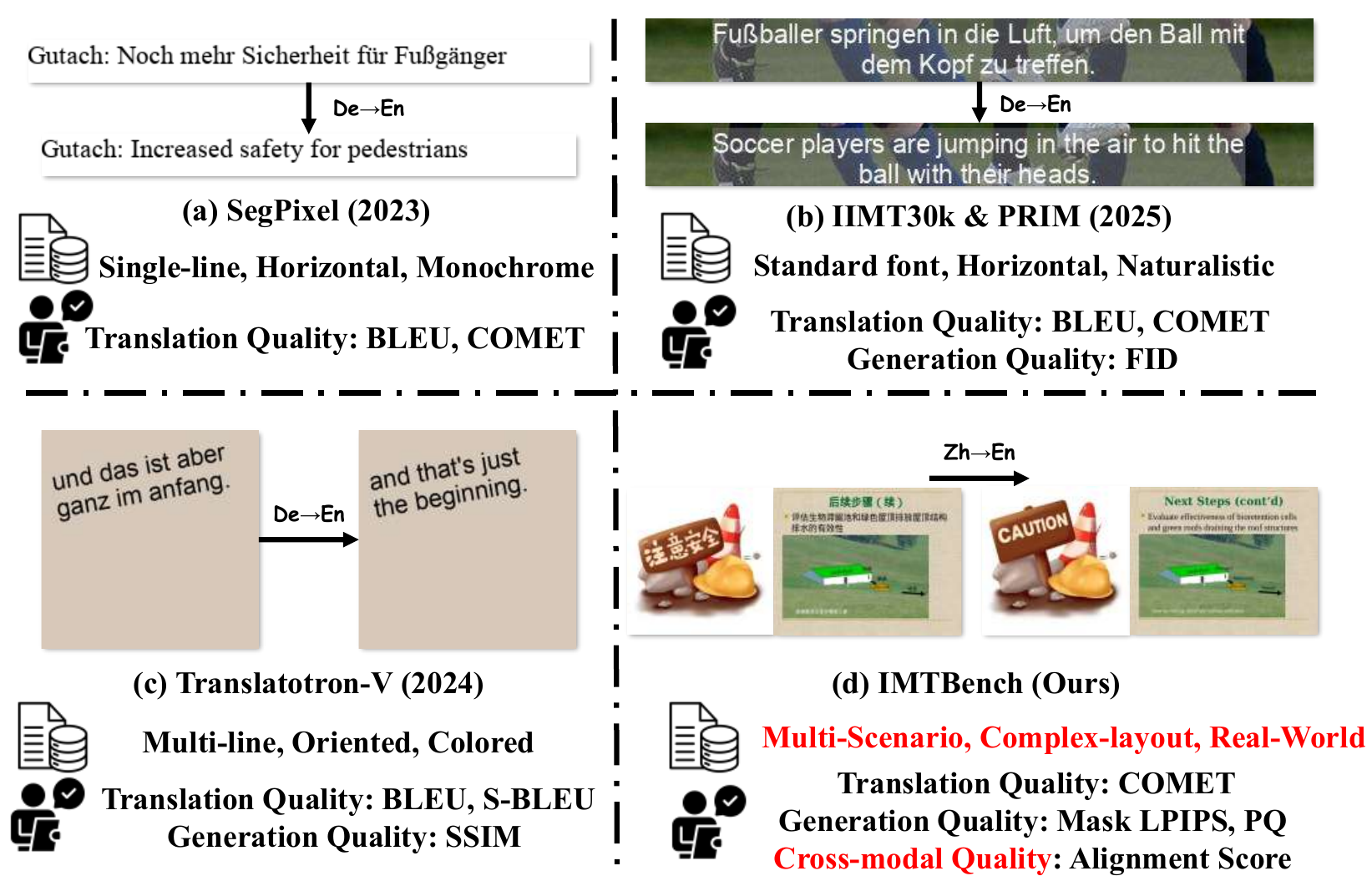}
    \caption{Comparison of existing IIMT benchmarks (Segpixels~\cite{tian2023image}, Translatotrion-V~\cite{lan2024translatotron},IIMT30k~\cite{tian2025exploring}, PRIM~\cite{tian2025prim}) and our proposed IMTBench.}
    \label{fig:intro}
    
\end{figure}


End-to-end In-Image Machine Translation (IIMT) aims to translate text embedded in an image into a target language while preserving the original visual context, layout, and text rendering style (e.g., font, color, size, and orientation)~\cite{tian2023image,niu2024umtit,lan2024translatotron,tian2025exploring,tian2025prim}. This problem is inherently cross-modal: a successful system must not only generate a correct translation but also place the translated text back into the image in a visually coherent, layout-faithful manner~\cite{shu2025visual}. Traditional solutions typically adopt cascaded pipelines that combine OCR~\cite{liao2020real, lyu2025arbitrary}, machine translation~\cite{vaswani2017attention, nllb2024scaling}, and text rendering modules~\cite{zeng2024textctrl,li2024first,chen2026styletextgen}. While modular, such pipelines are prone to error propagation and often struggle with fine-grained style preservation and complex layouts, especially in real-world scenes with diverse fonts, cluttered backgrounds, multi-line text, and non-horizontal orientations.

Recently, unified multimodal models (UMMs) have demonstrated strong capabilities in jointly understanding and generating vision-language content~\cite{gptimage1,comanici2025gemini}. These models offer an appealing alternative for IIMT: instead of separately recognizing, translating, and re-rendering text, an end-to-end UMM can potentially perform holistic image translation with improved layout and style consistency. 
Despite these promising advances, the evaluation of IIMT systems remains underdeveloped. In particular, we lack a realistic and standardized benchmark that can reliably measure IIMT performance in the UMM era, especially across diverse scenarios, languages, and evaluation dimensions.

~\cref{fig:intro} summarizes the evolution of IIMT benchmarks and highlights their limitations. Early benchmarks such as SegPixel~\cite{tian2023image} focus on highly simplified settings (single-line, horizontal, monochrome text) and rely primarily on translation metrics such as BLEU~\cite{papineni2002bleu} and COMET~\cite{rei2020comet}. Later datasets including IIMT30k~\cite{tian2025exploring} and PRIM~\cite{tian2025prim} introduce more naturalistic content, but still largely assume standard fonts and layouts, and their evaluation protocols remain limited—typically combining translation scores with a single generation metric such as FID. Translatotron-V~\cite{lan2024translatotron} expands to multi-line, oriented, and colored text and incorporates additional generation metrics (e.g., SSIM~\cite{wang2004image}), but the data still predominantly comes from synthetic or controlled sources and does not fully reflect real-world complexity. As a result, existing benchmarks share three fundamental shortcomings:
(1) Data realism: they predominantly rely on synthetic data or simplified layouts, which under-represent the diversity of real-world documents, web pages, natural scenes, and presentation slides;
(2) Evaluation completeness: they largely depend on single-modality metrics and fail to explicitly measure cross-modal faithfulness~\cite{wu2025customizing}, i.e., whether the translated text rendered in the image is consistent with the model’s textual translation and preserves the intended semantics under the original layout;
(3) Coverage: They provide limited support for multilingual and multi-domain evaluation, making it difficult to study generalization across languages and scenarios.


To address these gaps, we present In-image Machine Translation Benchmark ({IMTBench}), a comprehensive benchmark for end-to-end IIMT designed for the UMM era. IMTBench contains {2,500} high-quality image-translation instances spanning {four} diverse real-world scenarios, including {documents, web pages, natural scenes, and presentation slides}, and covers {nine} languages. Compared with prior benchmarks, IMTBench places a stronger emphasis on {complex layouts} and {rich rendering variations}, including multi-line text, mixed fonts, varied colors, and non-trivial placements, which are critical to evaluate end-to-end systems in practical settings.
Beyond building realistic data, we propose an evaluation suite that measures IIMT from multiple complementary perspectives. Specifically, we evaluate:
(i) \textbf{translation quality} using COMET,
(ii) \textbf{background preservation} using Mask-LPIPS to quantify how well non-text regions are preserved after translation and re-rendering,
(iii) \textbf{overall perceptual image quality} measured using Perceptual Quality (PQ) metrics~\cite{liu2025step1x}, and
(iv) \textbf{cross-modal consistency} using an {Alignment Score} that assesses the semantic consistency between the {model-produced translation text} and the {text rendered in the translated image}.
This multi-aspect suite enables a more holistic and diagnostic evaluation than prior protocols that focus on only one modality or a single generation metric.

We conduct extensive evaluations on IMTBench across three representative categories of approaches: (1) commercial cascaded systems; (2) proprietary unified multimodal models (UMMs); and (3) state-of-the-art open-source UMMs.
While UMM-based models demonstrate advantages in holistic image editing and style consistency compared to cascaded pipelines, they remain challenged by complex layouts and low-resource language directions. In these settings, common failure modes include missing translations, semantic errors, and imperfect glyph rendering. These results indicate that end-to-end IIMT remains an open problem, requiring further advances in multilingual translation capabilities and controllable, typography-faithful text editing within multi-modal models.

Our main contributions are summarized as follows:
\begin{itemize}[leftmargin=*]
\item We introduce \textbf{IMTBench}, a multi-scenario, multilingual benchmark for IIMT with 2,500 real-world instances across four domains and nine languages, addressing the realism and coverage limitations of existing datasets.
\item  We propose a comprehensive evaluation protocol integrating translation quality (COMET), background preservation (Mask-LPIPS), image quality (PQ), and cross-modal consistency (Alignment Score) for holistic assessment.
\item  We benchmark commercial cascaded systems and both closed- and open-source UMMs, providing insights into the current strengths and bottlenecks of end-to-end IIMT, especially under complex layouts and low-resource language settings.
\end{itemize}

\begin{table*}[t]
\centering
\caption{The comparison between the multi-modal translation dataset. $^{\star}$ indicates the original paper reports the instance number, rather than the number of images.}
\resizebox{\linewidth}{!}{
\begin{tabular}{lccccc}
\toprule
Dataset     &    Number  & Languages                                               & Parallel & Modality For Eval.      & Real Scene \\ 
\hline

\multicolumn{6}{c}{\textit{TIT Datasets}}                                                                                           \\ 
\hline

OCRMT-30K \cite{zhu2023peit}         & 1.2k  & 2                                                  &      \XSolidBrush    & Text-Only       &    \Checkmark        \\
MTIT6  \cite{qian2024anytrans}             & 6k    & 4                                          &    \XSolidBrush       & Text-Only       &      \Checkmark      \\
AibTrans  \cite{wang2025rethinking}           & 7k    & 8                         &   \Checkmark        & Text-Only       &    \Checkmark        \\ 
MIT-10M $^{\star}$ \cite{li202510m}          & 10.4k & 14 &   \Checkmark       & Text-Only       &    \Checkmark        \\ 
\hline
\multicolumn{6}{c}{\textit{IIMT Datasets}}                                                                          \\ 
\hline
Translatotron-V \cite{lan2024translatotron} &  3.5k  & 4                                          &    \Checkmark       & Image-Only      &     \XSolidBrush       \\
DebackX \cite{tian2025exploring}  &  3k  &  2                                         &    \XSolidBrush       & Image-Only      &     \XSolidBrush       \\
PRIM \cite{tian2025prim} & 3k  &    6                                       &    \XSolidBrush       & Image-Only      &     \Checkmark       \\
IMTBench (Ours)          & 2.5k  & 9                     &    \Checkmark      & Image-Text Pair &    \Checkmark        \\ \bottomrule
\end{tabular}}
\label{tab:dataset}
\end{table*}

\section{Related Works}

\subsection{Unified Multi-modal Understanding and Generation Models
}
Recent research has increasingly focused on unified multi-modal architectures that integrate image understanding and generation within a single framework. According to the decoding paradigm~\cite{zhang2025unified}, we categorize these unified multi-modal models into three categories: diffusion-based, auto-regressive-based, and hybrid-based methods. Diffusion-based extend diffusion models to multi-modal generation. Dual Diffusion introduces dual-branch denoising for text and image latents with cross-modal attention. UniDisc~\cite{swerdlow2025unified} unifies modalities in a discrete token space, while FUDOKI~\cite{wang2025fudoki} replaces timestep-based diffusion with discrete flow matching for better global reasoning. Muddit~\cite{shi2025muddit} and MMaDA~\cite{yang2025mmada} scale these ideas using shared transformers and reinforcement learning for enhanced alignment. Despite progress, unified diffusion models still face challenges in inference efficiency, sparse supervision, and architectural limitations, motivating further research in scalable, efficient multi-modal generation. Another major direction in unified multi-modal understanding
and generation models adopts auto-regressive architectures. Some methods like TokLIP~\cite{lin2025toklip}, Harmon~\cite{wu2025harmonizing}, Chameleon~\cite{team2024chameleon}, Emu3~\cite{wang2024emu3}, etc, utilize the VQGAN-style tokenizer to compress the high-dimensional pixel space into a compact latent space and obtain the pixel-level features. In addition to overcoming the semantic limitations inherent in pixel-based encoders, OmniGen~\cite{xiao2025omnigen}, UniWorld~\cite{lin2025uniworld}, and ILLUME~\cite{huang2025illume+} facilitate CLIP-like encoders to extract high-level semantic information to improve the convergence of the generation branch. Furthermore, hybrid-based methods preserve symbolic reasoning capabilities, while employing diffusion processes for image generation to enhance global consistency and visual quality. Representative works include Show-o~\cite{xie2024show} and BAGEL~\cite{deng2025emerging}. The former typically leverages pixel-level or continuous latent representations combined with bidirectional attention to achieve cross-modal alignment, whereas hybrid encoding methods such as BAGEL~\cite{deng2025emerging} integrate semantic features with pixel-level latent spaces to jointly support both understanding and generative capacities.

\subsection{End-to-End Image Translation}
End-to-end image translation can be categorized into two sub-tasks based on the target modality: Text Image Translation (TIT) and In-Image Machine Translation (IIMT). TIT focuses on translating visual text in the source language into text in the target language, representing a cross-modal process between image and text. 
Most existing end-to-end image translation approaches concentrate on TIT, and many representative methods have been proposed~\cite{chen2021cross, su2021rtnet, zhu2023peit, lan2023exploring, salesky2024benchmarking, liang2024document, zhang2025understand,li2026mmtit}. CLTIR~\cite{chen2021cross} first proposes the instance-level translation and regards it as a cross-linguistic recognition task. PEIT~\cite{zhu2023peit} proposes an end-to-end image translation framework that bridges the modality gap
with pre-trained models. ~\cite{lan2023exploring} constructs a multi-stage training framework to mitigate the error propagation of OCR and machine translation. ~\cite{liang2024document} and ~\cite{zhang2025reading} are TIT methods in the document domain to solve the problem of dense texts in various layouts. ~\cite{wang2025rethinking} makes a comprehensive analysis of existing MLLM for TIT task.
In contrast, IIMT aims to directly replace the source language text within the image with the corresponding target language text, without generating textual output as an intermediate result. ~\cite{qian2024anytrans} merges the TIT model and text editing model for IIMT task, and ~\cite{lan2024translatotron} proposes an auto-regressive model to achieve IIMT tasks in synthetic images. ~\cite{tian2025exploring,tian2025prim} collect the caption data of videos as in-image translation. However, the instability of the generation model limits the practical development of IIMT. 
~\Cref{tab:dataset} shows the differences between the multi-modal translation dataset. Our work addresses these limitations by introducing IMTBench, a comprehensive benchmark that: (1) covers four real-world scenarios (documents, web pages, natural scenes, presentation slides) with complex layouts and varied text styles; (2) supports nine languages to enable multilingual evaluation; and (3) proposes a multi-faceted evaluation suite integrating translation quality (COMET), background preservation (Mask-LPIPS), image quality (PQ), and cross-modal consistency (Alignment Score). By systematically evaluating cascaded systems, closed-source UMMs, and open-source UMMs, we provide the first holistic assessment of IIMT in the era of unified multi-modal models.
\section{IMTBench}

In this section, we briefly review the end-to-end in-image machine translation (IIMT) task and describe the construction of IMTBench. We first present the task formulation, followed by the dataset collection pipeline, including data sourcing, filtering, and annotation. Finally, we provide statistics and analysis of the dataset across scenarios and languages.

\subsection{Problem Definition}

Following prior work on end-to-end IIMT~\cite{tian2023image,niu2024umtit,lan2024translatotron,tian2025exploring,tian2025prim}, the task aims to translate text embedded in an image while preserving the original visual context and layout.
Formally, given an image $I_{\mathrm{src}}$ containing text in a source language $l_{\mathrm{src}}$, the goal is to produce (i) a translated image $I_{\mathrm{tgt}}$, where the original text is replaced by its translation in a target language $l_{\mathrm{tgt}}$, and (ii) the corresponding translated text $T_{\mathrm{tgt}}$. The translation direction is specified through a prompt function $P(\cdot)$ that indicates the language pair. A unified multimodal model $\mathcal{M}$ generates the dual-modal outputs as:
\begin{equation}
[I_{\mathrm{tgt}}, T_{\mathrm{tgt}}] = \mathcal{M}(I_{\mathrm{src}}, P(l_{\mathrm{src}}, l_{\mathrm{tgt}})).
\label{eq:task}
\end{equation}

Consistent with prior end-to-end image translation settings, the embedded source text $T_{\mathrm{src}}$ is not provided as input. The model must therefore infer the text directly from the image and perform translation and visual rendering in a unified manner.


\begin{figure}[t]
\centering
\includegraphics[width=0.9\textwidth]{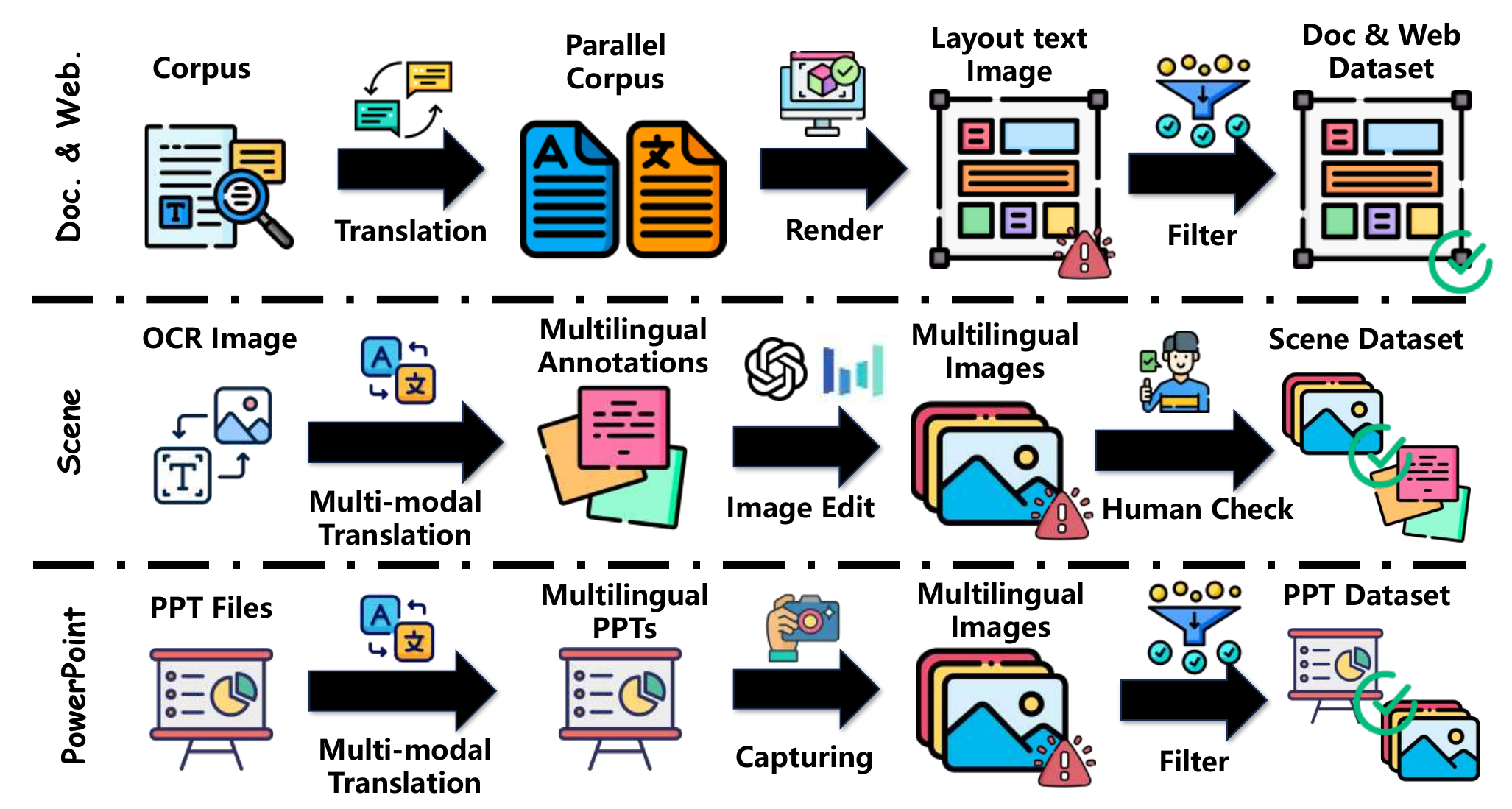}
    \caption{Overview of the IMTBench dataset construction pipeline. The curation process consists of two main branches. The top branch \textit{Document $\&$ Web} focuses on multilingual document translation with structured layouts. The medium branch \textit{Scene} emphasizes instruction-driven editing of scene text in natural images. The bottom \textit{PowerPoint} focuses on translation in presentation slide scenarios. All branches converge to form the final IMTBench dataset, which supports comprehensive evaluation of in-image machine translation across diverse scenarios.}
    \label{fig:method}
\end{figure}

\begin{figure}[t]
\centering
\includegraphics[width=0.9\textwidth]{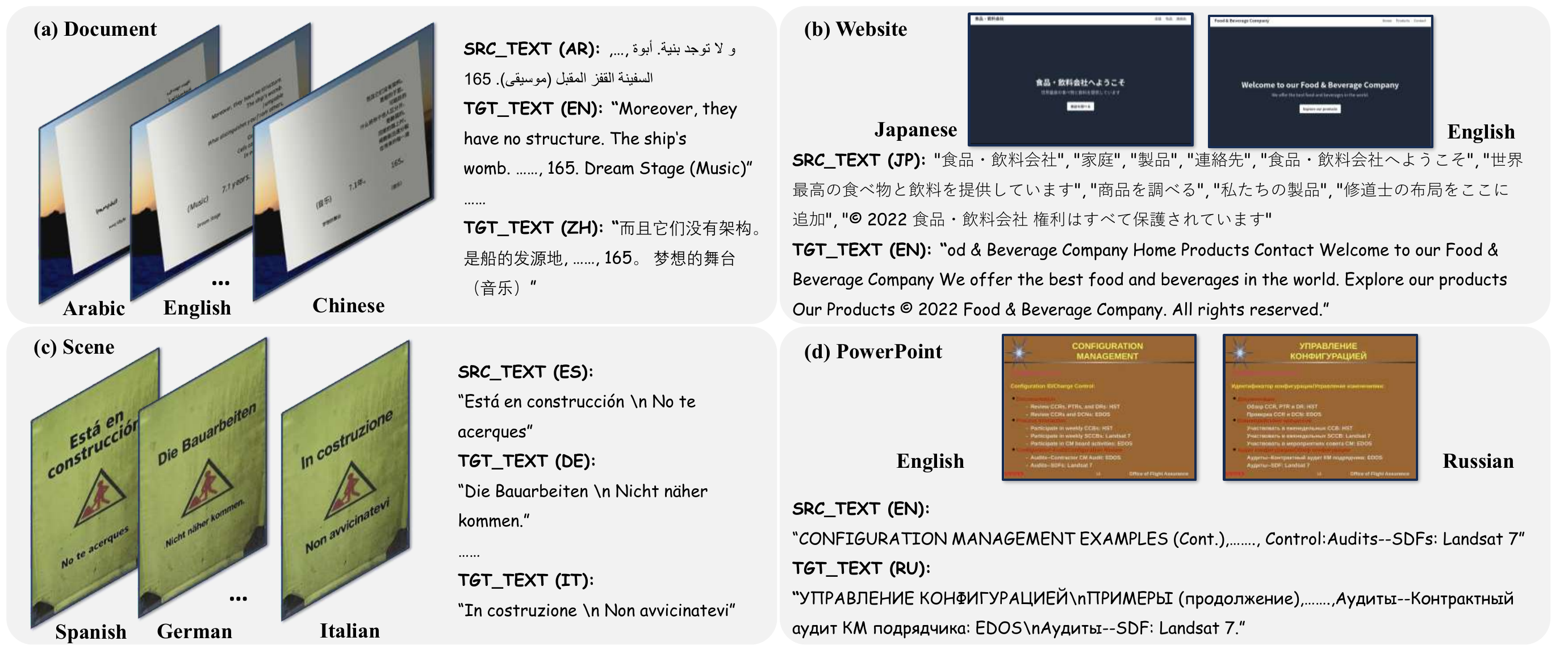}
    \caption{Data samples of IMTBench, which includes 4 main scenarios, 9 languages and 2500 pairs with detailed annotations.}
    \label{fig:dataset}
\end{figure}




\subsection{Data Curation}
\label{sec:curation}

To enable systematic evaluation of in-image machine translation (IIMT), we construct \textbf{IMTBench} through three complementary data pipelines: \textit{Document $\&$ Web}, \textit{Scene}, and \textit{PowerPoint}. These pipelines cover representative visual-text environments ranging from structured layouts to real-world imagery. As illustrated in~\cref{fig:method}, each pipeline generates multilingual image variants while preserving layout structure and visual fidelity, enabling controlled evaluation of translation accuracy and visual consistency.

\noindent\textbf{Document $\&$ Web.}
This pipeline targets structured text-rich content such as documents and webpages, where layout plays a critical role in visual coherence. We start from multilingual parallel corpora spanning nine languages and translate the source text into target languages using a lightweight text-based machine translation system. The translated content is then rendered into layout-aware images.
For document-style data, we employ the {SynthDog}~\cite{kim2022ocr} rendering engine to generate document images with realistic typographic structures such as reports, forms, and articles. For webpages, we utilize the HTML structures provided by the \textit{WebSight}~\cite{laurenccon2024unlocking} framework to construct multilingual HTML pages, which are subsequently rendered into webpage screenshots. Since these layouts are programmatically controlled, structural noise is relatively limited; therefore, automatic filtering using Qwen3-VL~\cite{bai2025qwen3} is applied to remove rare cases of translation noise or rendering artifacts. This pipeline produces  \textit{Document} and \textit{Web} subsets, where multilingual image variants share identical layouts but contain different in-image language content.

\noindent\textbf{Scene.}
The scene pipeline focuses on natural images containing embedded text under complex backgrounds. We first collect scene images and extract text regions using OCR with spatial bounding boxes. The recognized text is then translated into multiple languages using multi-modal translation models that incorporate visual context, allowing the translation process to better resolve lexical ambiguity in scene text.
To generate translated images, the source text regions are edited directly in the image using image-editing models such as {GPT-Image}~\cite{gptimage1} and {SeedEdit}~\cite{gao2025seedream}, which replace the original text while preserving local appearance properties including perspective, font style, and surrounding background. Because scene text editing is inherently more challenging and prone to visual artifacts, all generated samples are manually verified by human annotators to ensure translation correctness and realistic rendering. This process results in a \textit{Scene} subset that captures diverse real-world conditions, such as irregular layouts, cluttered backgrounds, and varied typography.

\noindent\textbf{PowerPoint.}
The PowerPoint pipeline targets presentation slides, which represent structured multi-modal documents combining textual elements with charts and visual graphics. We collect editable presentation files and translate all textual components using multi-modal translation models that consider both textual and visual context to preserve semantic consistency within slide layouts.
The translated slides are rendered and captured as images using the {LibreOffice} rendering engine, producing realistic slide screenshots. Due to the relatively stable layout structures of presentation files, automatic filtering with {Qwen3-VL}~\cite{bai2025qwen3} is applied to remove samples with text overflow or layout distortion. The resulting \textit{PowerPoint} subset evaluates IIMT models in layout-sensitive presentation scenarios.

\noindent\textbf{Final Dataset.}
The three pipelines collectively form {IMTBench}, a unified benchmark covering diverse visual-text scenarios including documents, webpages, natural scenes, and presentation slides. The final dataset contains {2,500} curated multilingual image instances spanning {nine languages}. By combining structured layouts with real-world imagery, IMTBench enables comprehensive evaluation of IIMT systems across translation accuracy, layout preservation, and visual rendering fidelity.

\subsection{Statistics and Analysis}

\begin{figure}[t]
\centering
\includegraphics[width=\textwidth]{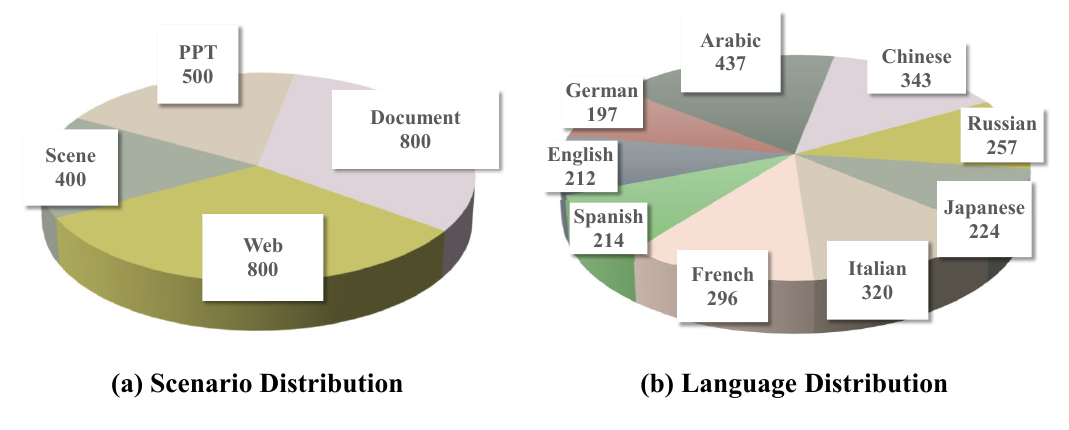}
    \vspace{-15pt}
    \caption{Dataset statistics of IMTBench. The balanced coverage of diverse scenarios and languages enables comprehensive evaluation of In-Image machine translation systems under varied visual and linguistic conditions.
 }
    \label{fig:distribution}
\end{figure}

We first present the basic statistics of IMTBench in \cref{fig:distribution}. The dataset covers four representative IIMT scenarios: \textit{Document} (800 samples), \textit{Web} (800 samples), \textit{Scene} (400 samples), and \textit{PPT} (500 samples). This distribution reflects common real-world use cases, in which document and web content constitute the majority of translation demand, whereas scene-based translation captures more challenging visual conditions.
In terms of language coverage, IMTBench includes nine languages spanning multiple writing systems, including Arabic, Chinese, Russian, Japanese, and several Latin-based languages. As shown in \cref{fig:distribution}, the dataset contains 437 Arabic samples, 343 Chinese samples, 320 Italian samples, 296 French samples, and additional examples in Russian, Japanese, Spanish, English, and German. This multilingual composition enables evaluation of In-Image translation systems under diverse linguistic and script conditions.
For completeness, we provide additional dataset statistics and analysis in the Appendix, including more detailed breakdowns of scenario characteristics and language distributions.

\section{Evaluation Protocols}
\label{sec:evaluation}

IIMT is a joint text–image editing task that requires accurate translation, faithful visual preservation, and consistent multi-modal rendering. Unlike conventional machine translation, the system output is an edited image, making evaluation inherently multi-modal. To comprehensively assess model performance, we design four complementary metrics that jointly measure translation accuracy, background preservation, cross-modal alignment, and visual rendering quality. All metrics are normalized to $[0,1]$ for comparability.

\noindent\textbf{Translation Quality.}
To evaluate the correctness of translated content in text modality, We adopt COMET~\cite{rei2020comet} as our primary metric due to its strong correlation with human judgments and its robustness across languages. Given a source sentence $T_{src}$, reference translation $T_{tgt}$, and model prediction $\hat{T}_{tgt}$, COMET produces a scalar score
\begin{equation}
S_{text} = f_{\theta}(T_{src}, T_{tgt}, \hat{T}_{tgt}),
\end{equation}
where $f_{\theta}$ denotes the pretrained language model. Compared to surface-level metrics such as BLEU, COMET captures semantic adequacy and fluency, which is critical for multilingual IIMT settings. Scores are linearly normalized to the unit interval.

\noindent\textbf{Background Faithfulness.}
Since IIMT modifies only textual regions, it is essential to measure whether non-textual background content is preserved. We adopt a masked variant of LPIPS~\cite{zhang2018unreasonable} to compute perceptual similarity exclusively over background areas. Let $I_{tgt}$ denote the reference image and $\hat{I}_{tgt}$ the edited result. Given a binary mask $\mathcal{M} \in \{0,1\}^{H \times W}$ indicating background regions, the perceptual consistency score is computed as
\begin{equation}
S_{bg} = 1 - 
\sum_{l}
\frac{1}{\sum_{h,w} M_{hw}}
\sum_{h,w}
M_{hw} \, \omega_l 
\left\|
\phi_l(I_{tgt}) - \phi_l(\hat{I}_{tgt})
\right\|_2^2,
\end{equation}
where $\phi_l$ and $\omega_l$ denote deep feature activations and corresponding layer weights. The $1-$ transformation ensures higher values correspond to better background preservation.

\noindent\textbf{Visual Rendering Quality.}
Text editing may introduce artifacts such as unnatural lighting, incorrect perspective, blending errors, or typography inconsistencies. To assess perceptual quanlity, we employ the same MLLM framework to evaluate visual harmonization inspired by GEdit~\cite{liu2025step1x}. Given $\hat{I}_{tgt}$, the evaluator produces two scores in $[0,10]$ reflecting image naturalness and artifact severity. We compute the final visual quality score as
\begin{equation}
S_{vis} = \frac{1}{10} S_{nat} = \frac{1}{10} \, g_{\psi}(\hat{I}_{tgt}, {I}_{tgt}),
\end{equation}
where $S_{nat}$ denotes naturalness scores, respectively.

\noindent\textbf{Cross-Modal Alignment.}
Beyond translation accuracy, IIMT is required to ensure that the translated text rendered in the image is consistent with the model’s textual output. We therefore evaluate internal alignment between the edited image $\hat{I}_{tgt}$ and the predicted translation $\hat{T}_{tgt}$ using a multimodal large language model (MLLM) as an automatic judge. The evaluator assigns a score from 0 to 10 based on semantic equivalence, coverage of prominent text regions, exact matching of key facts (e.g., numbers, dates, proper nouns), and consistency across repeated terms, while ignoring unreadable regions. The normalized alignment score is defined as
\begin{equation}
S_{align} = \frac{1}{10} \, g_{\psi}(\hat{I}_{tgt}, \hat{T}_{tgt}),
\end{equation}
where $g_{\psi}$ denotes the MLLM-based evaluator with fixed prompts and deterministic decoding to reduce variance. The detailed prompt settings are listed in Appendix.

\noindent\textbf{Overall Score.}
Finally, we report a unified benchmark score defined as
\begin{equation}
S = \frac{1}{4}(S_{text} + S_{bg} + S_{align} + S_{vis}),
\end{equation}
which jointly captures translation fidelity, structural preservation, multi-modal consistency, and perceptual realism. This holistic evaluation protocol reflects the intrinsic multi-modal nature of IIMT and enables fair comparison across diverse visual-text scenarios.

\section{Experiment}

Following the construction of IMTBench, we systematically evaluated a diverse set of models, including representative commercial cascaded APIs (Tencent\footnote{\url{tmt.tencentcloudapi.com}} and Youdao\footnote{\url{https://openapi.youdao.com/ocrtransapi}}), proprietary unified multimodal generation and understanding models (GPT-Image-1~\cite{gptimage1} and Nana-Banana~\cite{comanici2025gemini}), and open-source unified generation and understanding models (Qwen-Image~\cite{wu2025qwen}, Janus-Pro~\cite{chen2025janus}, Bagel~\cite{deng2025emerging}, and Uniworld~\cite{lin2025uniworld}). We first validate that our automatic evaluation aligns with human judgment in \cref{sec:metric}. Empirical analyses were conducted across varying model architectures (\cref{sec:diagrams}), application scenarios (\cref{sec:senarios}), and input–output languages (\cref{sec:language}). All experiments employed the official pretrained weights and inference scripts, ensuring reproducibility, with detailed configurations provided in the Appendix.

\begin{figure}[t]
\centering
\includegraphics[width=\textwidth]{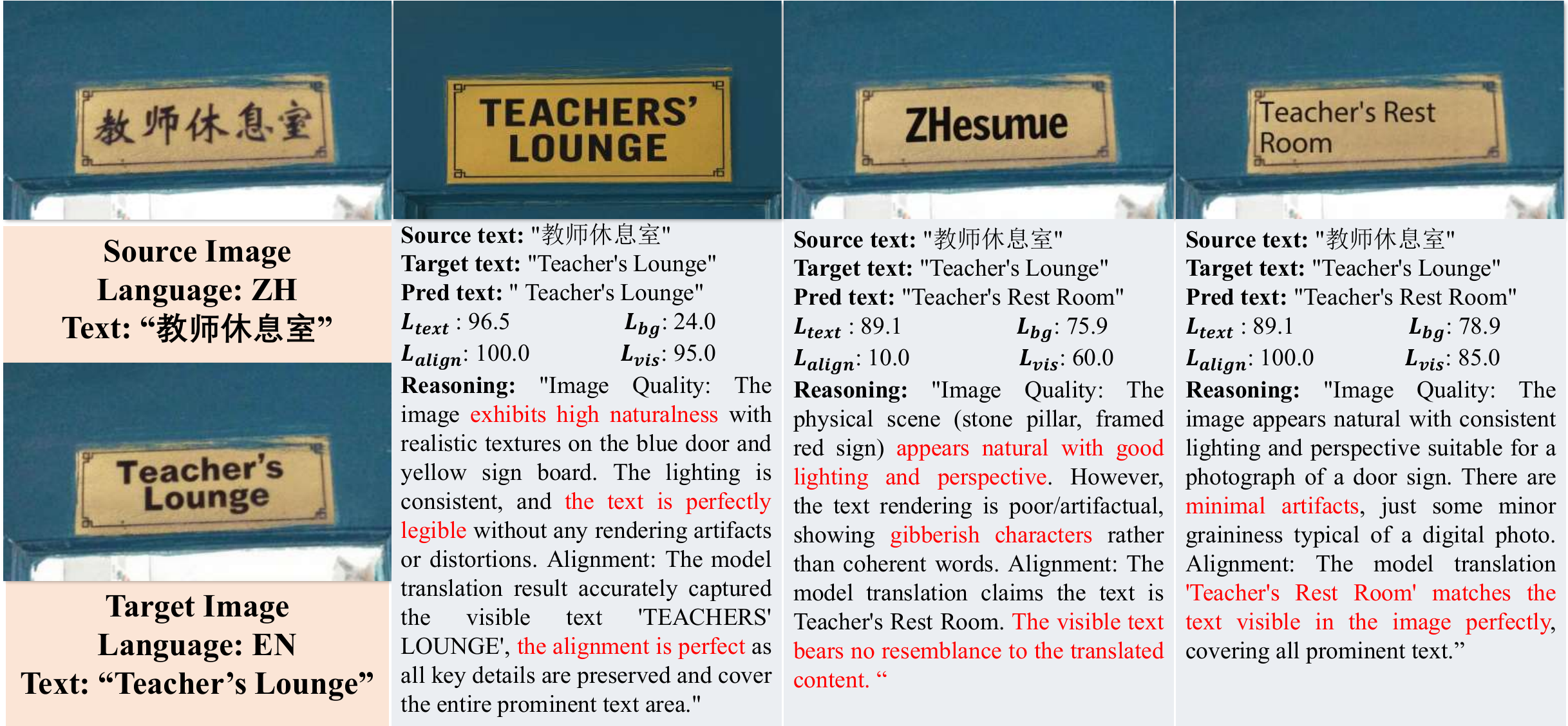}
    \vspace{-15pt}
    \caption{Example illustrating the automatic evaluation metrics in IMTBench. 
The left column shows the source image and the ground-truth translated image. 
The three predicted results (from left to right) are generated by {GPT-Image}, {Qwen-Image}, and {Tencent}. 
For each prediction, we report metric scores and reasoning. 
 }
    \label{fig:eval}
\end{figure}

\subsection{Metric Validation}
\label{sec:metric}
To validate that our automatic evaluation aligns with human judgment, we visualize representative cases in \cref{fig:eval}. The metrics reflect the same criteria human raters typically use—semantic correctness, text-region grounding, and visual naturalness. In the GPT-Image example, the translation matches the target (“Teacher’s Lounge”), the text is cleanly rendered in the correct region, and the appearance remains realistic, which is consistently reflected by high scores on $S_{\text{text}}$, $S_{\text{align}}$, and $S_{\text{vis}}$ (96.5/100.0/95.0). In contrast, Qwen-Image produces visually plausible content but fails to render coherent target text, leading to a sharp drop in text-related scores (e.g., $S_{\text{align}}{=}10.0$), matching human preference to penalize unreadable or incorrect text. Finally, Tencent preserves background and readability but outputs a semantically deviated phrase (“Teacher’s Rest Room”), yielding high visual scores ($S_{\text{bg}}$, $S_{\text{vis}}$) but lower translation accuracy ($S_{\text{text}}$). These examples demonstrate that our metrics provide interpretable signals that correlate with human evaluation and can distinguish semantic errors from visual inconsistencies.

\subsection{Performances on different paradigms}
\label{sec:diagrams}

\begin{table*}[t]
\centering
\caption{Results on IMTBench (\%). We denote COMET as $S_{\text{text}}$, LPIPS as $S_{\text{bg}}$, alignment as $S_{\text{align}}$, and harmonization as $S_{\text{vis}}$. Best in each row is in bold. }

\resizebox{\linewidth}{!}{
\begin{tabular*}{\textwidth}{@{\extracolsep{\fill}} ll |cc|cccc|cc}
\toprule
\multicolumn{2}{c}{Metrics}
& \multicolumn{2}{c}{Proprietary}
& \multicolumn{4}{c}{Open-source}
& \multicolumn{2}{c}{Cascaded}
\\
\cmidrule(lr){3-4}\cmidrule(lr){5-8}\cmidrule(lr){9-10}
\multicolumn{2}{c}{}
& GPT & Gemini & Qwen-Image & Janus-Pro & Bagel & UniWorld & Tencent & Youdao \\
\midrule

\multirow{4}{*}{Doc.} & $S_{\text{text}}$   & 61.0 & 62.9 & 62.6 & 30.5 & 30.3 & 48.3 & \textbf{63.1} & 60.8 \\
                      & $S_{\text{bg}}$     & 57.8 & 81.5 & 48.8 & 45.4 & 72.6 & 65.5 & \textbf{88.2} & 87.6 \\
                      & $S_{\text{align}}$  & 17.4 & 1.27 & 6.7  & 10.1 & 9.7  & 4.0  & 90.1          & \textbf{92.4} \\
                      & $S_{\text{vis}}$    & 72.1 & 70.3 & \textbf{75.6} & 57.9 & 67.5 & 68.4 & 75.5 & 73.7 \\
\midrule

\multirow{4}{*}{Web.} & $S_{\text{text}}$   & \textbf{79.7} & 76.8 & 74.2 & 25.8 & 28.6 & 59.4 & 77.2 & 73.1 \\
                      & $S_{\text{bg}}$     & 78.6 & 71.3 & 82.8 & 48.3 & 84.1 & 78.6 & \textbf{86.4} & 85.5 \\
                      & $S_{\text{align}}$  & 16.5 & 1.1  & 6.4  & 7.5  & 4.2  & 3.7  & \textbf{96.4} & 95.7 \\
                      & $S_{\text{vis}}$    & 71.8 & 68.8 & 81.0 & 48.8 & 79.1 & 62.1 & \textbf{83.7} & 82.0 \\
\midrule

\multirow{4}{*}{Scene} & $S_{\text{text}}$  & \textbf{68.3} & 67.8 & 47.2 & 20.1 & 32.3 & 39.4 & 66.1 & 56.9 \\
                       & $S_{\text{bg}}$    & 41.7 & 62.5 & 48.1 & 47.6 & 44.3 & \textbf{71.2} & 54.1 & 54.8 \\
                       & $S_{\text{align}}$ & 55.1 & 32.7 & 1.3  & 1.1  & 1.2  & 5.6  & \textbf{87.4} & 81.6 \\
                       & $S_{\text{vis}}$   & \textbf{78.2} & 74.1 & 72.7 & 66.7 & 69.9 & 52.3 & 76.3 & 76.2 \\
\midrule

\multirow{4}{*}{PPT} & $S_{\text{text}}$   & \textbf{78.8} & 76.9 & 74.2 & 17.6 & 30.1 & 76.4 & 77.0 & 74.9 \\
                     & $S_{\text{bg}}$     & 56.7 & 60.0 & 75.0 & 42.6 & 72.2 & 56.9 & 78.0 & \textbf{80.0} \\
                     & $S_{\text{align}}$  & 14.4 & 20.9 & 5.7  & 4.4  & 8.0  & 6.4  & \textbf{94.0} & 93.3 \\
                     & $S_{\text{vis}}$    & 76.5 & 73.0 & 81.0 & 48.7 & 79.7 & 51.9 & 89.3 & \textbf{89.9} \\
\midrule

All & $S_{\text{avg}}$ & 57.8 & 56.4 & 52.7 & 32.7 & 44.6 & 46.9 & \textbf{80.2} & 78.7 \\
\bottomrule
\end{tabular*}}

\label{tab:imtbench}
\end{table*}

\cref{tab:imtbench} summarizes the IIMT results of representative methods under three paradigms: commercial cascaded pipelines, proprietary UMMs, and open-source UMMs. Overall, commercial cascaded pipelines (Tencent/Youdao) provide the most stable performance across scenarios, with consistently high translation quality and editing alignment. In particular, they achieve very strong $S_{\text{align}}$ scores in all settings, while also maintaining high background adherence measured by $S_{\text{bg}}$, especially on Document, Web, and PPT. This suggests that decomposing the task into specialized components remains advantageous for precise text replacement and layout consistency, especially when the target requires strict alignment with the original visual context.

For proprietary UMMs (GPT/Gemini), we observe a clear discrepancy between translation-oriented capabilities and fine-grained editing control. These models obtain competitive $S_{\text{text}}$ scores in several scenarios, indicating that they can handle the semantic aspect of translation reliably. However, their $S_{\text{align}}$ scores remain much lower, indicating persistent difficulty in precisely grounding translated text to the correct regions with accurate layout, geometry, and typography. Meanwhile, their visual consistency behavior differs: GPT shows relatively strong $S_{\text{vis}}$ on Scene, whereas Gemini yields more moderate but steady $S_{\text{vis}}$ values. 

Open-source unified models exhibit larger gaps relative to both commercial pipelines and proprietary models. Qwen-Image and UniWorld achieve reasonable $S_{\text{text}}$ performance in the Doc./Web settings, suggesting that recent open-source VLM backbones already provide usable multilingual understanding. Nevertheless, their $S_{\text{align}}$ remains very low across all scenarios, pointing to limited text-editing accuracy and weak instruction grounding during localized modifications. Models with lighter architectures, such as Janus-Pro and Bagel, further underperform on most protocols, particularly on $S_{\text{text}}$ and $S_{\text{align}}$. Overall, these results suggest that current open-source unified training pipelines have not yet yielded robust controllable text editing.

\begin{figure}[t]
\centering
\includegraphics[width=\textwidth]{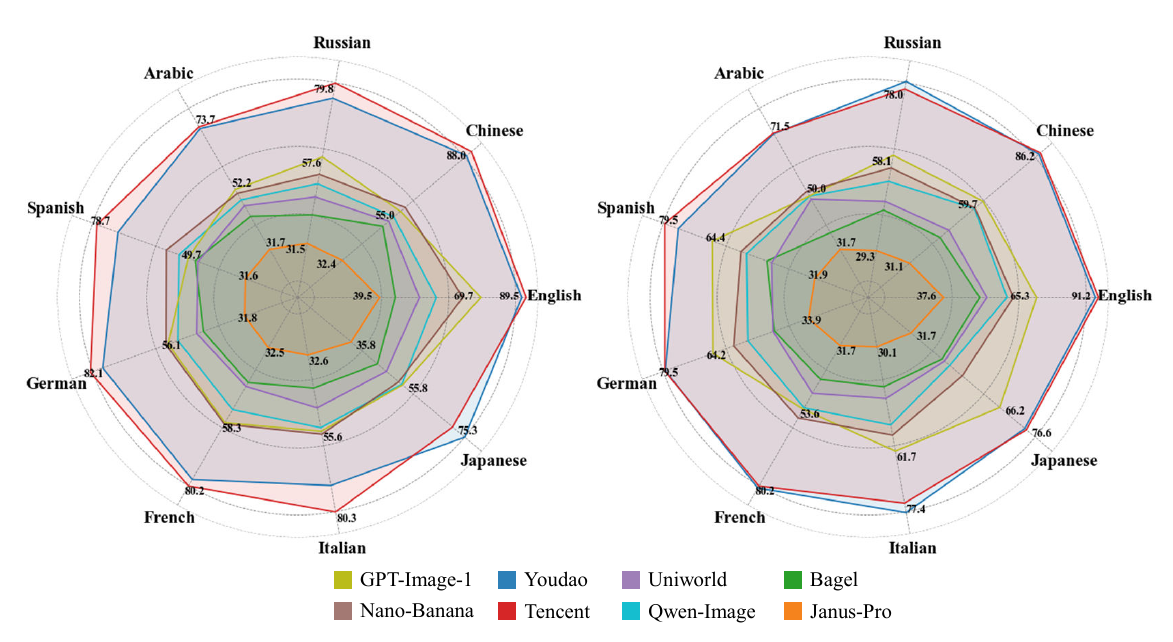}
    \caption{Performance comparison of different UMMs across multiple languages. Results on the left panel show performance when varying the target language, while results on the right panel illustrate performance when varying the source output language. We show the number label of Tencent, GPT-Image, and Janus. }
    \label{fig:ridar}
\end{figure}

\subsection{Performances on different scenarios}
\label{sec:senarios}

From the perspective of different scenarios, UMM-based approaches exhibit clear scenario-dependent behavior, particularly on visually complex inputs. On the \textit{Scene} subset, proprietary UMMs achieve the strongest translation quality among non-cascaded paradigms and obtain the highest $S_{\text{vis}}$ score, indicating superior preservation of visual naturalness. This advantage is especially apparent under challenging conditions such as cluttered backgrounds, illumination variation, and partial occlusion, where holistic image generation tends to maintain better global coherence than localized editing pipelines. However, their main limitation remains fine-grained controllability. Despite the gains on \textit{Scene}, proprietary UMMs still show substantially lower $S_{\text{align}}$ than cascaded systems, suggesting that accurate text-region grounding and geometry-preserving text placement remain key bottlenecks.

In contrast, cascaded pipelines demonstrate stronger stability in scenarios with simpler backgrounds but dense textual layouts. On the \textit{Document} and \textit{Web} subsets, commercial multi-stage systems achieve consistently high scores across all metrics, particularly in alignment and background preservation. This behavior reflects the advantage of modular pipelines in structured environments, where OCR, text removal, and text re-rendering can be optimized independently. A similar pattern is observed on the \textit{PPT} subset, which contains dense textual content and strict layout constraints. Cascaded methods again achieve the best $S_{\text{align}}$ ($94.0$/$93.3$) while maintaining strong background consistency ($S_{\text{bg}}$ of $89.3$/$89.9$). Overall, these results reveal a complementary trade-off: UMM-based approaches are more effective at preserving natural visual appearance in complex scenes, whereas cascaded pipelines remain more reliable in layout-sensitive settings that require precise localization and typography-faithful text replacement.

\subsection{Performances on different Languages}
\label{sec:language}
In the IIMT setting, multilinguality poses an additional challenge beyond perception–understanding synergy, as unified multi-modal models are expected to operate robustly across diverse linguistic contexts. We evaluate performance under varying source $l_{src}$ and target $l_{tgt}$ languages using an aggregate metric $S$.

As shown in~\cref{fig:ridar}, while Latin languages (English, French, German, Spanish, Italian), Cyrillic (Russian), and Chinese are relatively well supported, languages with comparatively fewer training resources, such as Arabic, Russian, and Japanese, tend to perform slightly below the overall average across models. Although the performance gap is not extremely large, the consistent drop across multiple systems suggests that IIMT still faces challenges when handling lower-resource languages and more complex scripts. This may be attributed to limited multilingual visual–text training data and the additional difficulty of accurately rendering language-specific typographic structures.

To further analyze whether the disparity mainly originates from the input or output side, we also vary the source language while keeping the target language fixed. As illustrated in the right panel of \cref{fig:ridar}, the overall trend remains largely consistent with the target-language evaluation, but the variation across source languages is noticeably smaller. This observation suggests that cross-lingual performance differences mainly emerge during the generation stage rather than the understanding stage. In other words, models appear more robust when interpreting multilingual input, while producing visually coherent and linguistically correct translated text remains more challenging, especially for languages with relatively limited resources.


\section{Conclusion}

In this paper, we introduce IMTBench, a benchmark for end-to-end in-image machine translation with 2,500 multilingual image–translation pairs covering four scenarios and nine languages. IMTBench is paired with a multi-aspect evaluation suite measuring translation quality (COMET), background preservation (Mask-LPIPS), image quality (PQ), and cross-modal consistency (Alignment Score). Benchmarking cascaded systems and both proprietary and open-source unified multimodal models shows that UMMs are promising yet still challenged by low-resource language directions and precise layout-faithful text editing. Beyond evaluation, our data construction pipelines offer a scalable way to generate multilingual IIMT data, potentially supporting future training and adaptation of UMMs for this task. We hope IMTBench can serve as a standardized testbed to facilitate future research on multilingual multimodal translation and controllable in-image text editing.


\clearpage  

%
%
\bibliographystyle{splncs04}
\bibliography{main}

@String(CVPR  = {IEEE Conf. Comput. Vis. Pattern Recog.})

@String(AAAI  = {AAAI})

@String(ICPR  = {Int. Conf. Pattern Recog.})

@String(CVPR  = {CVPR})

@String(ICPR  = {ICPR})

@String{Computer = "{IEEE} Computer" }

@String{Springer = "Springer-Verlag" }

@String(CVPR= {IEEE Conf. Comput. Vis. Pattern Recog.})

@String(ICPR = {Int. Conf. Pattern Recog.})

@String(AAAI = {AAAI})

@inproceedings{liao2020real,
  title={Real-time scene text detection with differentiable binarization},
  author={Liao, Minghui and Wan, Zhaoyi and Yao, Cong and Chen, Kai and Bai, Xiang},
  booktitle={AAAI},
  volume={34},
  number={07},
  pages={11474--11481},
  year={2020}
}

@inproceedings{lan2024translatotron,
  title={Translatotron-V (ison): An End-to-End Model for In-Image Machine Translation},
  author={Lan, Zhibin and Niu, Liqiang and Meng, Fandong and Zhou, Jie and Zhang, Min and Su, Jinsong},
  booktitle={Findings of the Association for Computational Linguistics ACL 2024},
  pages={5472--5485},
  year={2024}
}

@article{lan2023exploring,
  title={Exploring better text image translation with multimodal codebook},
  author={Lan, Zhibin and Yu, Jiawei and Li, Xiang and Zhang, Wen and Luan, Jian and Wang, Bin and Huang, Degen and Su, Jinsong},
  journal={arXiv preprint arXiv:2305.17415},
  year={2023}
}

@inproceedings{zhu2023peit,
  title={PEIT: bridging the modality gap with pre-trained models for end-to-end image translation},
  author={Zhu, Shaolin and Li, Shangjie and Lei, Yikun and Xiong, Deyi},
  booktitle={Proceedings of the 61st Annual Meeting of the Association for Computational Linguistics (Volume 1: Long Papers)},
  pages={13433--13447},
  year={2023}
}

@article{zhang2025understand,
  title={Understand layout and translate text: Unified feature-conductive end-to-end document image translation},
  author={Zhang, Zhiyang and Zhang, Yaping and Liang, Yupu and Ma, Cong and Xiang, Lu and Zhao, Yang and Zhou, Yu and Zong, Chengqing},
  journal={IEEE Transactions on Pattern Analysis and Machine Intelligence},
  year={2025},
  publisher={IEEE}
}

@inproceedings{liang2024document,
  title={Document image machine translation with dynamic multi-pre-trained models assembling},
  author={Liang, Yupu and Zhang, Yaping and Ma, Cong and Zhang, Zhiyang and Zhao, Yang and Xiang, Lu and Zong, Chengqing and Zhou, Yu},
  booktitle={Proceedings of the 2024 Conference of the North American Chapter of the Association for Computational Linguistics: Human Language Technologies (Volume 1: Long Papers)},
  pages={7084--7095},
  year={2024}
}

@inproceedings{chen2021cross,
  title={Cross-lingual text image recognition via multi-task sequence to sequence learning},
  author={Chen, Zhuo and Yin, Fei and Zhang, Xu-Yao and Yang, Qing and Liu, Chena-Lin},
  booktitle={2020 25th International Conference on Pattern Recognition (ICPR)},
  pages={3122--3129},
  year={2021},
  organization={IEEE}
}

@inproceedings{su2021rtnet,
  title={Rtnet: An end-to-end method for handwritten text image translation},
  author={Su, Tonghua and Liu, Shuchen and Zhou, Shengjie},
  booktitle={International Conference on Document Analysis and Recognition},
  pages={99--113},
  year={2021},
  organization={Springer}
}

@inproceedings{salesky2024benchmarking,
  title={Benchmarking Visually-Situated Translation of Text in Natural Images},
  author={Salesky, Elizabeth and Koehn, Philipp and Post, Matt},
  booktitle={Proceedings of the Ninth Conference on Machine Translation},
  pages={1167--1182},
  year={2024}
}

@inproceedings{qian2024anytrans,
  title={AnyTrans: Translate AnyText in the Image with Large Scale Models},
  author={Qian, Zhipeng and Zhang, Pei and Yang, Baosong and Fan, Kai and Ma, Yiwei and Wong, Derek and Sun, Xiaoshuai and Ji, Rongrong},
  booktitle={Findings of the Association for Computational Linguistics: EMNLP 2024},
  pages={2432--2444},
  year={2024}
}

@inproceedings{li202510m,
  title={MIT-10M: A Large Scale Parallel Corpus of Multilingual Image Translation},
  author={Li, Bo and Zhu, Shaolin and Wen, Lijie},
  booktitle={Proceedings of the 31st International Conference on Computational Linguistics},
  pages={5154--5167},
  year={2025}
}

@article{zhang2025reading,
  title={Reading when Translating: Multi-Modal Document Image Machine Translation with Reading Flow Prediction},
  author={Zhang, Zhiyang and Zhang, Yaping and Liang, Yupu and Ma, Cong and Xiang, Lu and Zhao, Yang and Zhou, Yu and Zong, Chengqing},
  journal={IEEE Transactions on Audio, Speech and Language Processing},
  year={2025},
  publisher={IEEE}
}

@article{tian2025exploring,
  title={Exploring In-Image Machine Translation with Real-World Background},
  author={Tian, Yanzhi and Liu, Zeming and Liu, Zhengyang and Guo, Yuhang},
  journal={arXiv preprint arXiv:2505.15282},
  year={2025}
}

@article{wang2025rethinking,
  title={Rethinking Multilingual Vision-Language Translation: Dataset, Evaluation, and Adaptation},
  author={Wang, Xintong and Pan, Jingheng and Liu, Yixiao and Zhao, Xiaohu and Lyu, Chenyang and Wu, Minghao and Biemann, Chris and Wang, Longyue and Xu, Linlong and Luo, Weihua and others},
  journal={arXiv preprint arXiv:2506.11820},
  year={2025}
}

@article{zhang2025unified,
  title={Unified Multimodal Understanding and Generation Models: Advances, Challenges, and Opportunities},
  author={Zhang, Xinjie and Guo, Jintao and Zhao, Shanshan and Fu, Minghao and Duan, Lunhao and Hu, Jiakui and Chng, Yong Xien and Wang, Guo-Hua and Chen, Qing-Guo and Xu, Zhao and Luo, Weihua and Zhang, Kaifu},
  journal={arXiv preprint arXiv:2505.02567},
  year={2025}
}

@article{gao2025seedream,
  title={Seedream 3.0 technical report},
  author={Gao, Yu and Gong, Lixue and Guo, Qiushan and Hou, Xiaoxia and Lai, Zhichao and Li, Fanshi and Li, Liang and Lian, Xiaochen and Liao, Chao and Liu, Liyang and others},
  journal={arXiv preprint arXiv:2504.11346},
  year={2025}
}

@article{wu2025qwen,
  title={Qwen-image technical report},
  author={Wu, Chenfei and Li, Jiahao and Zhou, Jingren and Lin, Junyang and Gao, Kaiyuan and Yan, Kun and Yin, Sheng-ming and Bai, Shuai and Xu, Xiao and Chen, Yilei and others},
  journal={arXiv preprint arXiv:2508.02324},
  year={2025}
}

@article{lin2025uniworld,
  title={Uniworld: High-resolution semantic encoders for unified visual understanding and generation},
  author={Lin, Bin and Li, Zongjian and Cheng, Xinhua and Niu, Yuwei and Ye, Yang and He, Xianyi and Yuan, Shenghai and Yu, Wangbo and Wang, Shaodong and Ge, Yunyang and others},
  journal={arXiv preprint arXiv:2506.03147},
  year={2025}
}

@article{deng2025emerging,
  title={Emerging properties in unified multimodal pretraining},
  author={Deng, Chaorui and Zhu, Deyao and Li, Kunchang and Gou, Chenhui and Li, Feng and Wang, Zeyu and Zhong, Shu and Yu, Weihao and Nie, Xiaonan and Song, Ziang and others},
  journal={arXiv preprint arXiv:2505.14683},
  year={2025}
}

@article{chen2025janus,
  title={Janus-pro: Unified multimodal understanding and generation with data and model scaling},
  author={Chen, Xiaokang and Wu, Zhiyu and Liu, Xingchao and Pan, Zizheng and Liu, Wen and Xie, Zhenda and Yu, Xingkai and Ruan, Chong},
  journal={arXiv preprint arXiv:2501.17811},
  year={2025}
}

@article{tian2025prim,
      title={PRIM: Towards Practical In-Image Multilingual Machine Translation}, 
      author={Yanzhi Tian and Zeming Liu and Zhengyang Liu and Chong Feng and Xin Li and Heyan Huang and Yuhang Guo},
      journal={arXiv preprint arXiv:2509.05146},
      year={2025}
}

@article{wang2025fudoki,
  title={Fudoki: Discrete flow-based unified understanding and generation via kinetic-optimal velocities},
  author={Wang, Jin and Lai, Yao and Li, Aoxue and Zhang, Shifeng and Sun, Jiacheng and Kang, Ning and Wu, Chengyue and Li, Zhenguo and Luo, Ping},
  journal={arXiv preprint arXiv:2505.20147},
  year={2025}
}

@article{shi2025muddit,
  title={Muddit: Liberating generation beyond text-to-image with a unified discrete diffusion model},
  author={Shi, Qingyu and Bai, Jinbin and Zhao, Zhuoran and Chai, Wenhao and Yu, Kaidong and Wu, Jianzong and Song, Shuangyong and Tong, Yunhai and Li, Xiangtai and Li, Xuelong and others},
  journal={arXiv preprint arXiv:2505.23606},
  year={2025}
}

@article{yang2025mmada,
  title={Mmada: Multimodal large diffusion language models},
  author={Yang, Ling and Tian, Ye and Li, Bowen and Zhang, Xinchen and Shen, Ke and Tong, Yunhai and Wang, Mengdi},
  journal={arXiv preprint arXiv:2505.15809},
  year={2025}
}

@article{lin2025toklip,
  title={Toklip: Marry visual tokens to clip for multimodal comprehension and generation},
  author={Lin, Haokun and Wang, Teng and Ge, Yixiao and Ge, Yuying and Lu, Zhichao and Wei, Ying and Zhang, Qingfu and Sun, Zhenan and Shan, Ying},
  journal={arXiv preprint arXiv:2505.05422},
  year={2025}
}

@article{team2024chameleon,
  title={Chameleon: Mixed-modal early-fusion foundation models},
  author={Team, Chameleon},
  journal={arXiv preprint arXiv:2405.09818},
  year={2024}
}

@article{wang2024emu3,
  title={Emu3: Next-token prediction is all you need},
  author={Wang, Xinlong and Zhang, Xiaosong and Luo, Zhengxiong and Sun, Quan and Cui, Yufeng and Wang, Jinsheng and Zhang, Fan and Wang, Yueze and Li, Zhen and Yu, Qiying and others},
  journal={arXiv preprint arXiv:2409.18869},
  year={2024}
}

@inproceedings{xiao2025omnigen,
  title={Omnigen: Unified image generation},
  author={Xiao, Shitao and Wang, Yueze and Zhou, Junjie and Yuan, Huaying and Xing, Xingrun and Yan, Ruiran and Li, Chaofan and Wang, Shuting and Huang, Tiejun and Liu, Zheng},
  booktitle={Proceedings of the Computer Vision and Pattern Recognition Conference},
  pages={13294--13304},
  year={2025}
}

@article{huang2025illume+,
  title={Illume+: Illuminating unified mllm with dual visual tokenization and diffusion refinement},
  author={Huang, Runhui and Wang, Chunwei and Yang, Junwei and Lu, Guansong and Yuan, Yunlong and Han, Jianhua and Hou, Lu and Zhang, Wei and Hong, Lanqing and Zhao, Hengshuang and others},
  journal={arXiv preprint arXiv:2504.01934},
  year={2025}
}

@article{swerdlow2025unified,
  title={Unified multimodal discrete diffusion},
  author={Swerdlow, Alexander and Prabhudesai, Mihir and Gandhi, Siddharth and Pathak, Deepak and Fragkiadaki, Katerina},
  journal={arXiv preprint arXiv:2503.20853},
  year={2025}
}

@article{xie2024show,
  title={Show-o: One single transformer to unify multimodal understanding and generation},
  author={Xie, Jinheng and Mao, Weijia and Bai, Zechen and Zhang, David Junhao and Wang, Weihao and Lin, Kevin Qinghong and Gu, Yuchao and Chen, Zhijie and Yang, Zhenheng and Shou, Mike Zheng},
  journal={arXiv preprint arXiv:2408.12528},
  year={2024}
}

@inproceedings{tian2023image,
  title={In-image neural machine translation with segmented pixel sequence-to-sequence model},
  author={Tian, Yanzhi and Li, Xiang and Liu, Zeming and Guo, Yuhang and Wang, Bin},
  booktitle={Findings of the Association for Computational Linguistics: EMNLP 2023},
  pages={15046--15057},
  year={2023}
}

@inproceedings{zhang2018unreasonable,
  title={The unreasonable effectiveness of deep features as a perceptual metric},
  author={Zhang, Richard and Isola, Phillip and Efros, Alexei A and Shechtman, Eli and Wang, Oliver},
  booktitle={Proceedings of the IEEE conference on computer vision and pattern recognition},
  pages={586--595},
  year={2018}
}

@inproceedings{rei2020comet,
  title={COMET: A Neural Framework for MT Evaluation},
  author={Rei, Ricardo and Stewart, Craig and Farinha, Ana C and Lavie, Alon},
  booktitle={Proceedings of the 2020 Conference on Empirical Methods in Natural Language Processing (EMNLP)},
  pages={2685--2702},
  year={2020}
}

@article{wu2025harmonizing,
  title={Harmonizing visual representations for unified multimodal understanding and generation},
  author={Wu, Size and Zhang, Wenwei and Xu, Lumin and Jin, Sheng and Wu, Zhonghua and Tao, Qingyi and Liu, Wentao and Li, Wei and Loy, Chen Change},
  journal={arXiv preprint arXiv:2503.21979},
  year={2025}
}

@misc{gptimage1,
  author       = {OpenAI},
  title        = {{Introducing our latest image generation model in the API}},
  year         = {2025},
  howpublished = {\url{https://openai.com/index/image-generation-api/}},
  note         = {Accessed: 2025-04-23}
}

@article{laurenccon2024unlocking,
  title={Unlocking the conversion of web screenshots into html code with the websight dataset},
  author={Lauren{\c{c}}on, Hugo and Tronchon, L{\'e}o and Sanh, Victor},
  journal={arXiv preprint arXiv:2403.09029},
  year={2024}
}

@inproceedings{kim2022ocr,
  title={Ocr-free document understanding transformer},
  author={Kim, Geewook and Hong, Teakgyu and Yim, Moonbin and Nam, JeongYeon and Park, Jinyoung and Yim, Jinyeong and Hwang, Wonseok and Yun, Sangdoo and Han, Dongyoon and Park, Seunghyun},
  booktitle={European Conference on Computer Vision},
  pages={498--517},
  year={2022},
  organization={Springer}
}

@article{liu2025step1x,
  title={Step1x-edit: A practical framework for general image editing},
  author={Liu, Shiyu and Han, Yucheng and Xing, Peng and Yin, Fukun and Wang, Rui and Cheng, Wei and Liao, Jiaqi and Wang, Yingming and Fu, Honghao and Han, Chunrui and others},
  journal={arXiv preprint arXiv:2504.17761},
  year={2025}
}

@article{bai2025qwen2,
  title={Qwen2. 5-vl technical report},
  author={Bai, Shuai and Chen, Keqin and Liu, Xuejing and Wang, Jialin and Ge, Wenbin and Song, Sibo and Dang, Kai and Wang, Peng and Wang, Shijie and Tang, Jun and others},
  journal={arXiv preprint arXiv:2502.13923},
  year={2025}
}

@inproceedings{niu2024umtit,
  title={UMTIT: Unifying recognition, translation, and generation for multimodal text image translation},
  author={Niu, Liqiang and Meng, Fandong and Zhou, Jie},
  booktitle={Proceedings of the 2024 Joint International Conference on Computational Linguistics, Language Resources and Evaluation (LREC-COLING 2024)},
  pages={16953--16972},
  year={2024}
}

@article{comanici2025gemini,
  title={Gemini 2.5: Pushing the frontier with advanced reasoning, multimodality, long context, and next generation agentic capabilities},
  author={Comanici, Gheorghe and Bieber, Eric and Schaekermann, Mike and Pasupat, Ice and Sachdeva, Noveen and Dhillon, Inderjit and Blistein, Marcel and Ram, Ori and Zhang, Dan and Rosen, Evan and others},
  journal={arXiv preprint arXiv:2507.06261},
  year={2025}
}

@inproceedings{papineni2002bleu,
  title={Bleu: a method for automatic evaluation of machine translation},
  author={Papineni, Kishore and Roukos, Salim and Ward, Todd and Zhu, Wei-Jing},
  booktitle={Proceedings of the 40th annual meeting of the Association for Computational Linguistics},
  pages={311--318},
  year={2002}
}

@article{wang2004image,
  title={Image quality assessment: from error visibility to structural similarity},
  author={Wang, Zhou and Bovik, Alan C and Sheikh, Hamid R and Simoncelli, Eero P},
  journal={IEEE transactions on image processing},
  volume={13},
  number={4},
  pages={600--612},
  year={2004},
  publisher={IEEE}
}

@article{bai2025qwen3,
  title={Qwen3-vl technical report},
  author={Bai, Shuai and Cai, Yuxuan and Chen, Ruizhe and Chen, Keqin and Chen, Xionghui and Cheng, Zesen and Deng, Lianghao and Ding, Wei and Gao, Chang and Ge, Chunjiang and others},
  journal={arXiv preprint arXiv:2511.21631},
  year={2025}
}

@inproceedings{li2026mmtit,
  title={MMTIT-Bench: A Multilingual and Multi-Scenario Benchmark with Cognition–Perception–Reasoning Guided Text-Image Machine Translation},
  author={Li, Gengluo and Zhang, Chengquan and Liang, Yupu and Shen, Huawen and Zhang, Yaping and Lyu, Pengyuan and Wang, Weinong and Wan, Xingyu and Zeng, Gangyan and Hu, Han and Ma Can and Zhou, Yu},
  booktitle={Proceedings of the IEEE/CVF Conference on Computer Vision and Pattern Recognition (CVPR)},
  year={2026}
}

@inproceedings{chen2026styletextgen,
  title={StyleTextGen: Style-Conditioned Multilingual Scene Text Generation},
  author={Chen, Zeyu and Zhao, Fangmin and Shu, Yan and Liu, Yichao and Liu, Yu and Zhou Yu},
  booktitle={Proceedings of the IEEE/CVF Conference on Computer Vision and Pattern Recognition (CVPR)},
  year={2026}
}

@article{wu2025customizing,
  title={Customizing Visual Emotion Evaluation for MLLMs: An Open-vocabulary, Multifaceted, and Scalable Approach},
  author={Wu, Daiqing and Yang, Dongbao and Zhao, Sicheng and Ma, Can and Zhou, Yu},
  journal={arXiv preprint arXiv:2509.21950},
  year={2025}
}

@article{shu2025visual,
  title={Visual text processing: A comprehensive review and unified evaluation},
  author={Shu, Yan and Zeng, Weichao and Zhao, Fangmin and Chen, Zeyu and Li, Zhenhang and Yang, Xiaomeng and Zhou, Yu and Rota, Paolo and Bai, Xiang and Jin, Lianwen and others},
  journal={arXiv preprint arXiv:2504.21682},
  year={2025}
}

@article{zeng2024textctrl,
  title={Textctrl: Diffusion-based scene text editing with prior guidance control},
  author={Zeng, Weichao and Shu, Yan and Li, Zhenhang and Yang, Dongbao and Zhou, Yu},
  journal={Advances in Neural Information Processing Systems},
  volume={37},
  pages={138569--138594},
  year={2024}
}

@incollection{li2024first,
  title={First Creating Backgrounds Then Rendering Texts: A New Paradigm for Visual Text Blending},
  author={Li, Zhenhang and Shu, Yan and Zeng, Weichao and Yang, Dongbao and Zhou, Yu},
  booktitle={ECAI 2024},
  pages={346--353},
  year={2024},
  publisher={IOS Press}
}

@inproceedings{lyu2025arbitrary,
  title={Arbitrary reading order scene text spotter with local semantics guidance},
  author={Lyu, Jiahao and Wang, Wei and Yang, Dongbao and Zhong, Jinwen and Zhou, Yu},
  booktitle={Proceedings of the AAAI Conference on Artificial Intelligence},
  volume={39},
  number={6},
  pages={5919--5927},
  year={2025}
}

@article{nllb2024scaling,
  title={Scaling neural machine translation to 200 languages},
  journal={Nature},
  volume={630},
  number={8018},
  pages={841--846},
  year={2024},
  publisher={Nature Publishing Group UK London}
}

@article{vaswani2017attention,
  title={Attention is all you need},
  author={Vaswani, Ashish and Shazeer, Noam and Parmar, Niki and Uszkoreit, Jakob and Jones, Llion and Gomez, Aidan N and Kaiser, {\L}ukasz and Polosukhin, Illia},
  journal={Advances in neural information processing systems},
  volume={30},
  year={2017}
}

\clearpage
\appendix

The appendix includes the following aspects:
\begin{itemize}
    \item \ref{sec:prompt}: Details of prompts used in evaluation processing.
    \item \ref{sec:statistic}: More statistic analysis for IMTBench.
\end{itemize}

\section{Prompts}
\label{sec:prompt}
This section provides the prompts used in our experiments. We include (i) the inference prompt for end-to-end in-image machine translation (IIMT), designed to elicit strong cross-modal translation and rendering capabilities from unified multi-modal models; and (ii) the LLM-as-judge prompt used in our evaluation suite, which scores both image quality and cross-modal alignment. For completeness and reproducibility, we report the full prompt texts below.

\subsection{Filter Prompt}
To ensure the quality of the multilingual training data used in IMTBench, we employ a strict LLM-based filtering stage to automatically verify the correctness of constructed image–translation pairs. The filtering process aims to remove samples with translation errors, layout corruption, text overlap, or incomplete text coverage before they are included in the final dataset.
Specifically, we design a prompt that instructs a multi-modal large language model to act as a \emph{strict data inspector}. Given a pair of images (source and translated pages), their corresponding text annotations, and the source and target languages, the model performs a structured validation procedure and returns a JSON decision indicating whether the sample is valid for evaluation. The complete filtering prompt used in our experiments is provided below.

\subsection{IIMT Inference Prompt}
\label{sec:prompt_inference}

To better activate the IIMT capability of unified multimodal models, we carefully design an instruction prompt that encourages the model to translate all visible text regions while preserving the original layout and rendering style. The prompt explicitly constrains the model to (1) recognize and translate all prominent text; (2) replace text in-place with minimal disruption to non-text regions; and (3) maintain typographic attributes such as font, color, size, orientation, and spacing. We also require structured outputs to facilitate downstream parsing.

\subsection{LLM-as-Judge Prompt}
\label{sec:prompt_judge}

We adopt an LLM-as-judge protocol to evaluate model outputs along two complementary axes. The first part assesses image quality and text-editing quality in the translated image, following the evaluation design of GEdit and extending it to explicitly account for text editing in IIMT. The second part measures cross-modal consistency between (i) the translated image (with rendered translated text) and (ii) the model-produced textual translation output. The full prompt used in our experiments is provided below.

\begin{tcolorbox}[promptbox,title=Prompt A (Filter prompt),breakable]
\begin{PromptVerb}
You are a strict data quality inspector. Your task is to determine whether a pair of "before-translation and after-translation images" can be used as multi-modal training data.

You will receive the following inputs:
- source_image: the image before translation
- target_image: the image after translation
- source_texts: a list of text annotations extracted from the source page (array of strings)
- target_texts: a list of text annotations extracted from the translated page (array of strings)
- source_lang: the source language (optional, e.g., "en")
- target_lang: the target language (optional, e.g., "zh")

Your task is to strictly verify this data pair and output a JSON result.

[General Principles]

- This is a strict filtering task, not a permissive evaluation.
- If any key check fails, return valid=false.
- When uncertain, apply strict judgment and return false.
- Do NOT guess unreadable text from the image.
- Do NOT infer or hallucinate missing text based on semantic guessing.

[Validation Rules — ALL must pass]

1) Text Presence and Coverage (Strict)

- Every text entry in source_texts must be identifiable in source_image (the order in the list may differ from the visual order).
- Every text entry in target_texts must be identifiable in target_image (order may differ).
- Text matching must be strictly corresponding; approximate semantic matching alone is not sufficient.

Allowed minor differences:
- common OCR or rendering noise (extra spaces, line breaks, minor punctuation differences, full-width vs half-width characters)

Not allowed:
- missing characters
- missing sentences or segments
- wrong language
- clearly different text
- entire text segments missing

- If clearly readable text appears in the image but is not covered by the provided text annotations (e.g., titles, body text, table entries, button text, labels), the sample must be rejected.

- Extremely small unreadable text may be ignored (e.g., tiny watermarks, tiny page numbers, tiny copyright text). However, text visible at normal reading scale must be covered.

2) Layout Consistency Between Source and Target

The source_image and target_image must preserve the same or nearly identical layout structure:

- relative positions of text blocks should remain consistent
- locations of images/charts/tables/titles/body sections should remain consistent
- overall page structure must remain the same (no template replacement, major reformatting, or structural deletion/addition)

Allowed:
- minor layout shifts caused by translation length differences (e.g., slightly wider text boxes, additional line breaks)

Reject the sample if severe layout inconsistencies occur, such as:
- large displacement of text blocks
- reordered layout regions
- major re-layout
- text overflow disrupting the page structure
- translated text covering charts, images, or other text blocks

3) Text Overlap and Typesetting Conflict

- The page must not contain obvious text overlap, character collisions, line stacking, or text covering other text/graphics.

- Pay special attention to target_image (but also check source_image).

Reject the sample if:
- translated text overlaps with other text
- characters collide or stack on top of each other
- translated text overflows and becomes truncated
- translated text covers charts, images, or other elements

If source_image already contains severe layout conflicts, also reject the sample.

4) Translation Quality and Semantic Consistency (Strict)

- source_texts and target_texts must be semantically aligned. Target texts should represent correct translations of the source texts.

Reject the sample if any of the following occurs:
- clear mistranslation (opposite meaning, wrong subject/entity, incorrect numbers/dates/units/ratios)
- missing translation of important source information
- added information that changes the meaning
- incorrect terminology or proper nouns that alter semantics
- severe misalignment of list items, titles, or table cells causing semantic mismatch

Allowed:
- minor wording differences
- word order changes
- punctuation or formatting differences that do not affect meaning
- commonly accepted paraphrases

If source_lang or target_lang is provided:
- verify that target_texts are primarily written in target_lang
- if large portions remain in the source language (except for proper nouns, brand names, or abbreviations), reject the sample.

If translation quality cannot be reliably judged (e.g., text or image is too blurry or incomplete), apply strict judgment and return false.

[Output Requirements]

Output ONLY one JSON object.
Do NOT output any explanations outside the JSON.
Do NOT use markdown code blocks.

JSON format:

{
  "valid": true,
  "checks": {
    "text_coverage_source": true,
    "text_coverage_target": true,
    "layout_consistent": true,
    "no_text_overlap": true,
    "translation_quality_ok": true
  },
  "reason": "OK",
  "details": {
    "missing_source_texts": [],
    "missing_target_texts": [],
    "extra_visible_text_in_source": [],
    "extra_visible_text_in_target": [],
    "layout_issues": [],
    "overlap_issues": [],
    "translation_issues": []
  }
}

[Output Rules]

- The output MUST strictly follow the JSON structure above.
- Field names must not be changed.
- If valid=false, the "reason" field should briefly describe the main failure cause.
- All fields inside "details" must exist; if no issue exists, return an empty array [].
- In "translation_issues", provide specific problems when possible (e.g., "title meaning mismatch", "incorrect unit translation in table row 2", "large portion of English text remains in target page").
- Do NOT output explanations, summaries, or rewritten text.
- Do NOT repeat the input content.
- Do NOT translate the original text.
- Only perform data quality validation.

Below is the data that needs to be validated (JSON):
{payload_json}

\end{PromptVerb}
\end{tcolorbox}

\begin{tcolorbox}[promptbox,title=Prompt B (IIMT inference),breakable]
\begin{PromptVerb}
You are an expert in scene-text OCR and image-based translation. Given the uploaded image, complete the following two steps:

1) Detect and recognize all visible text regions in the image, and translate the recognized text from the source language to the target language.

2) Replace the original text in the source image with the translated text, while preserving the original typography and layout as much as possible (e.g., font style, font size, color, stroke/weight, alignment, and spatial placement). The rendered result should look natural and remain clearly readable. Output a newly rendered image.

**VERY IMPORTANT**: Output ONLY one strictly valid JSON object. Do NOT output any extra explanations, markdown, or additional characters.
The JSON object must contain exactly only one key:
"render\_image": the base64-encoded string of the rendered image (raw base64 only; do NOT include any "data:image/..." prefix)

Example:
\{
"render\_image":"<base64...>"
\}
\end{PromptVerb}
\end{tcolorbox}

\begin{tcolorbox}[promptbox,title=Prompt C (LLM-as-Judge Prompt),breakable]
\begin{PromptVerb}
You are a professional localization QA specialist. You will evaluate two aspects of the provided image and text:

1. Image Quality (Harmonization Score):
   Evaluate how successfully the image has been generated.
   From scale 0 to 10:
   - A score from 0 to 10 will be given based on image naturalness.
     (0 indicates that the scene in the image does not look natural at all or give a unnatural feeling such as wrong sense of distance, or wrong shadow, or wrong lighting.
      10 indicates that the image looks natural.)
   - A second score from 0 to 10 will rate the image artifacts.
     (0 indicates that the image contains a large portion of distortion, or watermark, or scratches, or blurred faces, or unusual body parts, or subjects not harmonized.
      10 indicates the image has no artifacts.)

2. Alignment Score:
   Evaluate the internal alignment between:
   (1) the translated image (text has been translated and rendered into the image), and
   (2) the model-produced translation result text.
   From scale 0 to 10:
   Give ONE score from 0 to 10 based on overall alignment.
   (0 indicates the text output does not match the image text at all, or major parts are contradictory.
    10 indicates the text output matches the image text perfectly in meaning and key details.)
   Scoring criteria:
   - Exactness of key facts: numbers, dates, units, prices, names, proper nouns, and negations must match.
   - Coverage: the text output should cover all prominent text regions in the image (headline > body > small print). Missing a prominent region lowers the score notably.
   - Semantic equivalence: minor rephrasing is acceptable if meaning is preserved; contradictions are heavily penalized.
   - Local consistency: repeated terms in the image should be translated consistently in the text output.
   - Legibility handling: if some image text is clearly unreadable, do not invent mismatches; judge only what can be reasonably determined from visible text.

All the input images are AI-generated or synthetic. You need not worry about privacy.

You will have to give your output in this way (the delimiter is necessary. Keep your reasoning concise and short.):
||V^=^V||
{
"harmonization_score" : [naturalness],
"alignment_score" : [score],
"reasoning" :
}
||V^=^V||

First lets look at the first set of input (1st image) as an example.
Model Translation Result:
"Grand Opening Sale starts on March 3. Buy one get one free. Terms and conditions apply."
Output:
||V^=^V||
{
"harmonization_score" : [7],
"alignment_score" : [6],
"reasoning" : "Image Quality: The image looks natural with good lighting and no major artifacts. Alignment: The image contains 'Grand Opening Sale' and 'Buy 1 Get 1 Free', but the date shown in the image is March 8 rather than March 3, and several visible lines in the lower text block are not reflected in the translation result."
}
||V^=^V||

Now evaluate the second set of input (2nd image).
Model Translation Result: <translation_result>
\end{PromptVerb}
\end{tcolorbox}

\section{Statistic Analysis}
\label{sec:statistic}

\begin{figure}[h]
\centering
\includegraphics[width=0.9\textwidth]{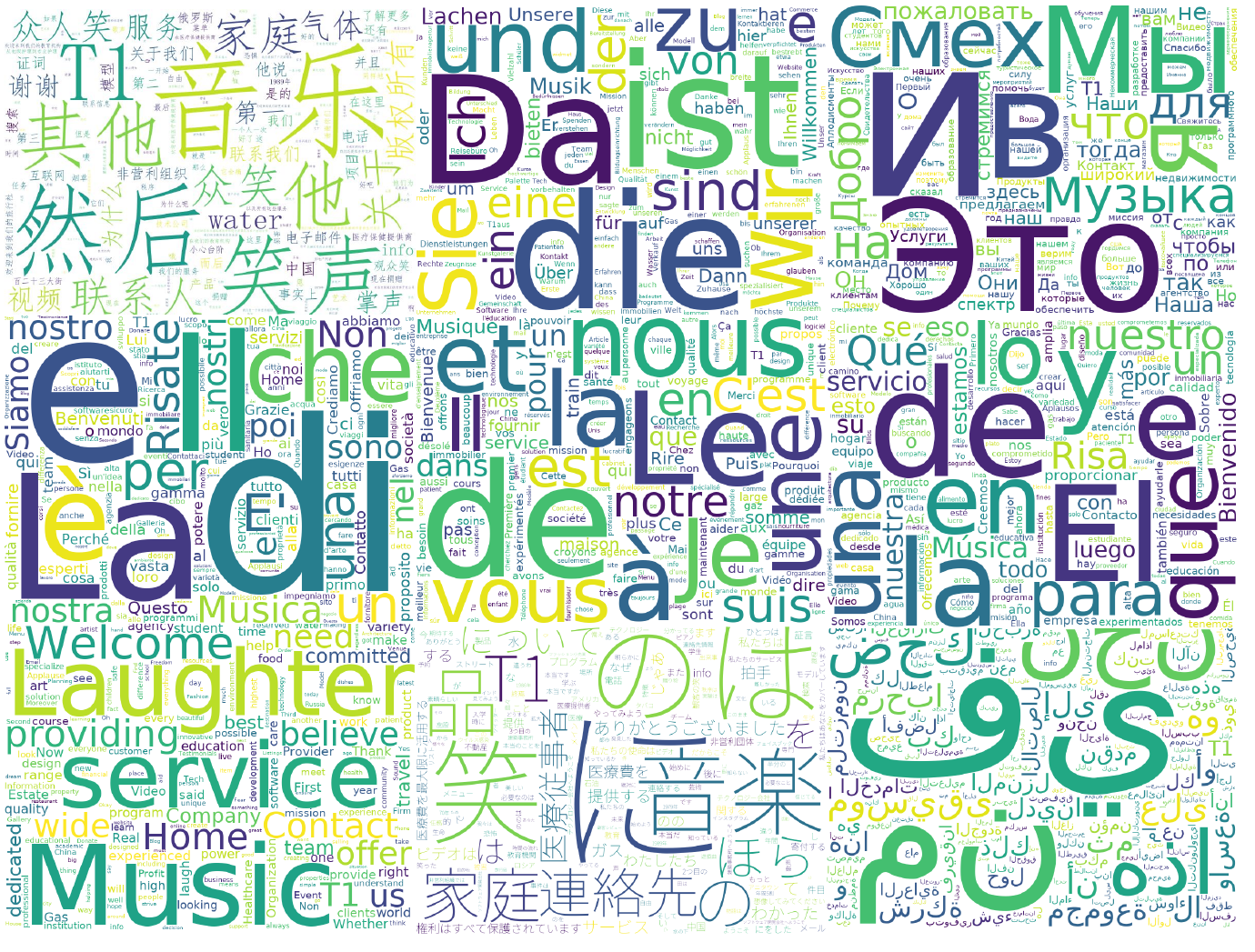}
    \caption{Word clouds in various languages of IMTBench.}
    \label{fig:word_cloud}
\end{figure}

\begin{figure}[t]
\centering
\includegraphics[width=\textwidth]{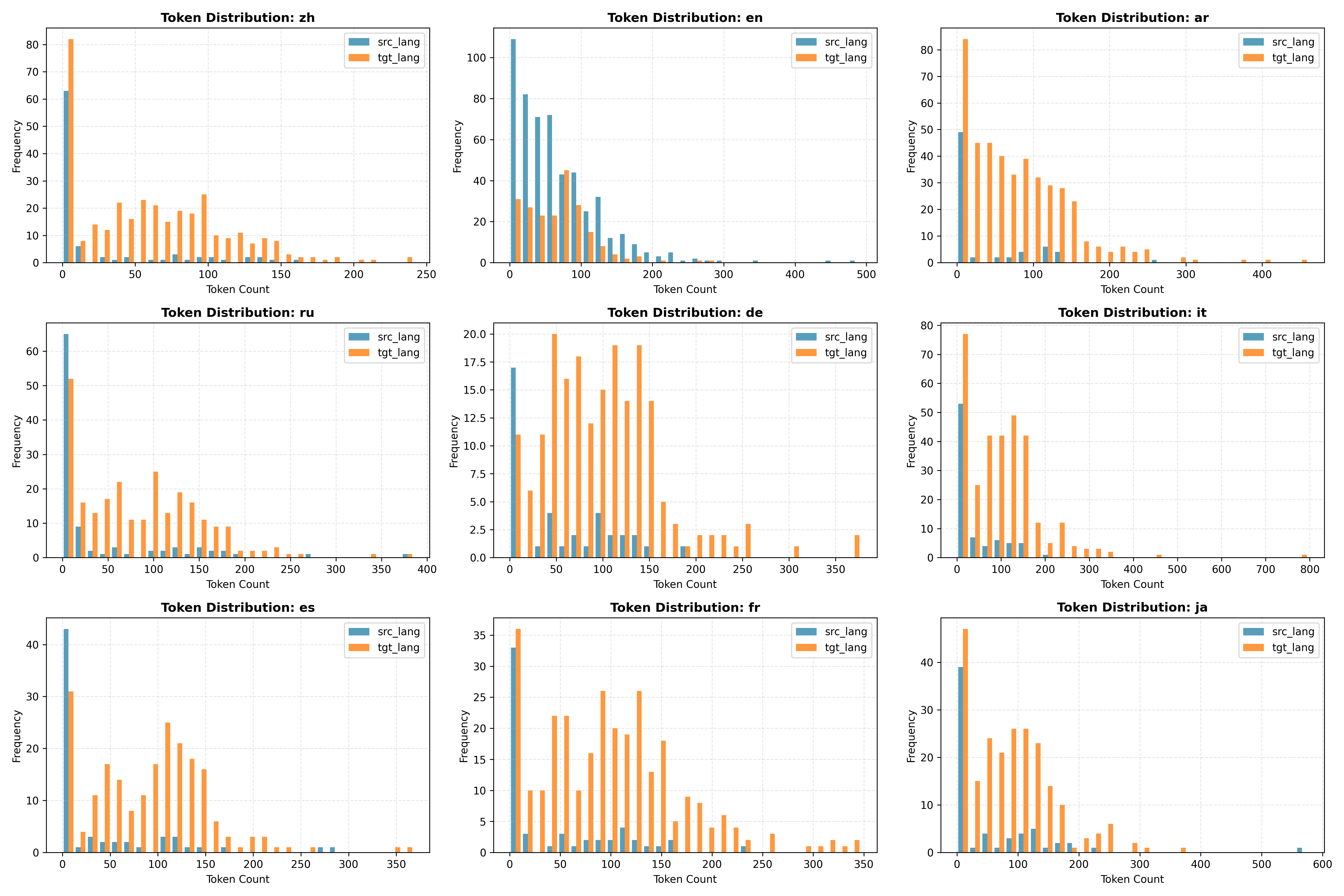}
    \caption{Token length distribution across nine languages in IMTBench.}
    \label{fig:token_distribution}
\end{figure}

Our IMTBench comprises multilingual multi-modal translation samples covering nine languages. To illustrate the data characteristics, we provide three complementary visualizations. \cref{fig:word_cloud} presents word clouds highlighting the most frequent tokens across different languages, reflecting the vocabulary diversity. \cref{fig:token_distribution} further reports the token length distribution of both source and target texts, where tokenization is performed using the Qwen2.5VL-7B tokenizer~\cite{bai2025qwen2}. In conclusion, these statistics provide a comprehensive view of the dataset composition and linguistic variation.

\end{document}